# Machine Learning Evaluation of the Echo-Chamber Effect in Medical Forums

Marina Sokolova
IBDA@Dalhousie University and University of Ottawa
sokolova@uottawa.ca

Victoria Bobicev
Technical University of Moldova
victoria.bobicev@ia.utm.md

**Abstract.** We propose the Echo-Chamber Effect assessment of an online forum. Sentiments perceived by the forum readers are at the core of the analysis; a complete message is the unit of the study. We build 14 models and apply those to represent discussions gathered from an online medical forum. We use four multi-class sentiment classification applications and two Machine Learning algorithms to evaluate prowess of the assessment models.

## 1. Echo-Chamber Effect

An echo chamber refers to discussions in which only certain ideas, information and beliefs are shared among the discussion participants. Definitions of an echo chamber emphasize participants' selective consumption of information (Garimella et al, 2018), avoidance of opposing information (Zollo et al, 2015), fragmentation of participants into narrow, homogenous groups (Colleoni et al, 2014; Shore et al, 2018). Markgraf and Schoch (2019) identify the following prerequisites of an echo chamber: Social Boundaries (i.e., discussions are limited within a narrow group), Information Homogeneity (i.e., interaction with information confirming previous believes), User Similarity (i.e., the position-conforming information is provided by like-minded individuals).

Two aspects of the Echo-Chamber Effect (ECE) studies have prompted our current research. First, significant majority of ECE research were focused on online political discussions (Dubois & Blank, 2018). At the same time, emergence of COVID-19 and the ensuing pandemics necessitated studies of ECE in medical online discussions. Those studies are imperative in prevention of misinformation, be it misinformation about vaccines, treatments, or spread of a disease (Lovari, 2020). Next, ECE is actively and predominantly studied in the Twitter discussions (Colleoni et al, 2014; Shore et al, 2018; Cossard et al, 2020) and large-scale social networks, e.g., Facebook, Reddit (Cinelli et al, 2020). Discussions happening on the other online platforms do not attract as much attention.

In the current study, we address Echo-Chamber Effect in discussions happening on an online medical forum. Our study is based on readers' perception of sentiments expressed by a message. Although we work with 80 discussions posted on the IVF Ages 35+ sub-forum[1], our lexicon-independent approach can be used in ECE studies in other medical forums. We present 14 models of the Echo-Chamber Effect (ECE) assessment. We use multi-class sentiment classification to evaluate the model prowess. In this study, *Precision (macro)* serves as the main criteria of the model performance.

Previously, we have analyzed transitions of sentiments among messages (Bobicev et al, 2015) and the impact of author's activity on recognition of sentiments (Sokolova & Bobicev, 2015; Bobicev & Sokolova, 2015). Our ECE assessment extends and generalizes those results.

---

[1] http://ivf.ca/forums/forum/166-ivf-ages-35/

## 2. Related Work on Echo-Chamber Effect and the Online Language

From perspective of Natural Language Processing and Text Analytics, studies of Echo-Chamber Effect in online forums closely relate to studies of online communications, including communication and social network studies, sentiment analysis, and studies of the online language phenomena (Fig 1).

The online language" is a third category recognized in present-day communication (Crystal 2006; Ehret &Taboada, 2020); spoken language and written language are the two primary categories of verbal communication (Biber et al, 1999). This section discusses communication concepts central to the ECE assessment in online forums.

A reader-centric approach is important for a text-based assessment of Echo-Chamber Effect. Readers' perception of the posted messages includes the subjective stance and opinions perceived from the message. Such perception reflects on readers' background, education, expertise, and level of interest in reading the text (Davoodi & Kosseim, 2017).

In online communications, articulation and acceptance of the emotional support depends on the communication skills and the emotional competence of interlocutors. Communication and media studies had shown that participation in online communicates helps to mitigate considerable psychosocial risks and highly unpredictable outcomes, frequently associated with treatments and health conditions (Van der Akker, 2015). Giving and receiving emotional support has positive effects on emotional well-being for patients with higher emotional communication. At the same time, highly emotional exchanges have detrimental impacts on emotional well-being for those with lower emotional communication competence (Yoo et al, 2014). For internet-based peer-support groups, distress before and after online chat and depressive symptoms have a strong negative correlation with digital literacy of participants (Lepore et al, 2019). Those conflicting tendencies can be exaggerated by Echo-Chamber Effect. In the online communities with a strong Echo-Chamber Effect, participants may ignore communities that support different opinions.

Social network analysis is essential to ECE studies. Two studies of the vaccination debates follow formation of echo-chambers on Twitter. In Italy, vaccination advocates avoid mentioning vaccination skeptics, thus increasing ECE in their networks (Cossard et al, 2020). In US, groups of antivaxxers and provaxxers showed a high homophily, a strong tendency for people to establish ties with people who are similar to themselves in socially significant ways (Mønsted & Lehmann, 2019).

Sentiment propagation and sentiment influence have been subjects of Sentiment Analysis, an advanced field of Natural Language Processing. High sentiment correlation was found in messages posted in online forum discussions (Weroński et al, 2012). Studies of hyperlink connections in blogs have shown that connections are strongly influenced by immediate posts but further influence steeply declines (Miller et al, 2011). Zafarani et al. (2010) studied the sentiment propagation in a LiveJournal dataset. They concluded that sentiment propagation in user's network positively correlates with the number of friends and negatively correlates with the number of posts and prolificacy of the user's friends. Persistence of positive and negative expressions may considerably vary: rapidly-fading information contains significantly more words related to negative emotions, whereas the persistent information contains more words related to positive emotions (Wu et al, 2011; Hansen et al, 2011).

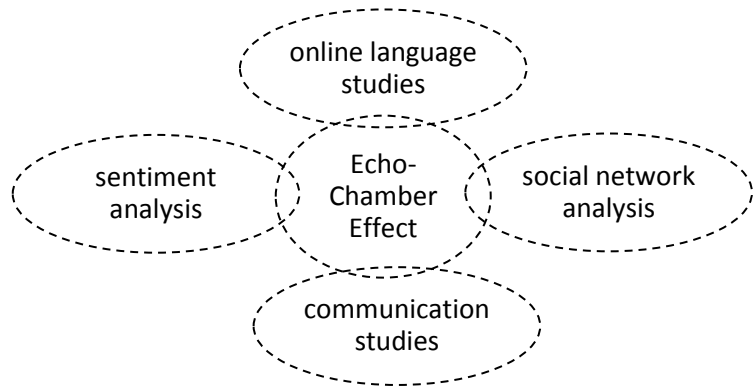

Figure 1: Components of the Echo-Chamber Effect studies – NLP perspective.

## 3. Echo-Chamber Effect Models

3.1 Parameters

We propose ECE models that use three categories of message parameters: i) sentiments perceived by the message annotators; ii) authors' activity in the discussion and on the forum; and iii) the message posting in the discussion (Fig 2). Noticeably, our models do not use the textual content of the messages.

Preliminary studies of the data identified peer-to-peer support, patient empowerment, as well as patient's uncertainty as the major reasons for the discussions' participations. Manual annotation of the IVF data set resulted in 5 sentiment categories: 'confusion', 'encouragement', 'gratitude', 'factual', and 'endorsement', a transitional category between 'factual' and 'encouragement'. Annotators used the reader perception model; (Sokolova & Bobicev, 2013) provide details of the annotation process of the IVF data. Those sentiments cover, albeit not exclusively, complex psychological factors of infertility (Hocaoglu, 2018).

For the current study, each post was annotated by two independent raters using pre-defined categories 'confusion', 'encouragement', 'gratitude', 'factual', and 'endorsement'. The annotators reached a strong agreement with Fleiss Kappa = 0.737. The posts assigned two different categories were considered 'ambiguous'.

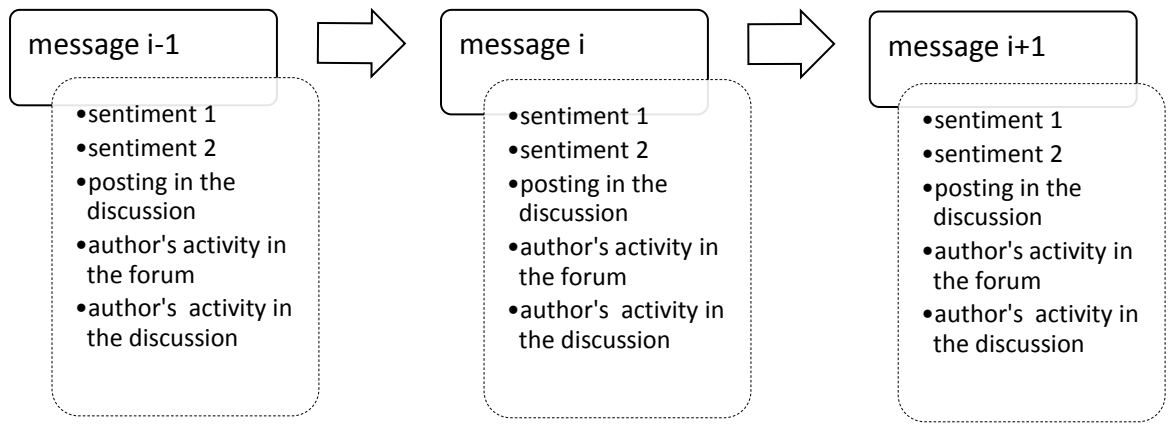

**Figure 2: Factors in the Echo-Chamber Effect estimate.**

We looked at three factors of author's activity on the forum:
1) initiations of new discussions;
2) the total number of messages posted by an author;
3) contribution to discussions initiated by other authors.

The authors who start discussions (aka first authors) actively participated in the initiated discussion and guided it in the direction they needed. Only in 10% of discussions, the first authors did not participate in discussion after initiating it. On average, 25% of messages per discussion were posted by the author of the first post. We identified a group of more active, or prolific, authors who posted approximately 7 times more posts that an average author. We estimated the "prolificity" of the authors as the ratio of the total number of author's posts to the total number of posts of the most prolific author in the studied data (Patil et al, 2013). Thus, prolificity ranges between [0, 1] and the participant with the greatest number of posts has prolificity equal to 1. In our data, the average prolificity of the prolific authors is 0.44, while the overall prolificity is 0.06. We hypothesize that sentiments of messages posted by an interlocutor already involved in discussion can be easier identifiable than sentiments in a message posted by a person just joining this thread. Thus, we introduced a category of discussion newcomers, i.e. authors contributing to discussion for the first time.

Our ECE models factor in the message posting within discussion, as we hypothesize that postings can affect readers' perception of the message sentiment. For example, readers may not strongly perceive sentiments expressed at the beginning of discussion and then better "tune" to sentiments as discussion progresses. [ In (Sokolova & Bobicev, 2015), we reported that annotators assigned different sentiment to 26% of the first posts and to 16% of the last post, whereas only 13% of all the posts were labeled ambiguous, i.e. were annotated with different sentiments.] Our current models test this hypothesis by differentiating among three types of postings: beginning ($1^{st}$ post), end (the last post), and middle of discussion (all the other posts).

3.2 The ECE models

Our first goal was to demonstrate that there are patterns of sentiments in forum's discussions and they mutually influence each other. Hence, we built a representation which reflected sentiment transitions in discussions. Having two annotation labels for each post we decided to use them both as parameters rather than merge them. This allowed us to disambiguate the *ambiguous* label, which appeared when two annotators selected different sentiment labels for the post:

- **Model I** – 4 categorical parameters. We represented each post through the two labels assigned by each annotator to the previous post and two labels assigned by each annotator to the following post.
For first and last posts, we used parameters "none" as a proxy of absent labels.
- **Model II** – 4 categorical parameters + 3 binary parameters = 7 parameters, where the four categorical parameters are the annotation labels (Model I), and the three binary parameters show whether the previous, current and next messages are first, middle, or last ones.
- **Model III** – 4 categorical parameters + 2 binary parameters + 1 numerical parameter = 7 parameters, where the four categorical parameters are the annotation labels (Model I) and the three other parameters represent author's activity (author's parameters): a binary parameter *f* indicating whether the author of the post is the one who started this discussion; a binary parameter *n* indicating whether the author posted in this discussion for the first time; a numerical parameter *pr* containing authors "prolificity", calculated as described in Section 4.2. Note that these parameters are independent and can simultaneously be true.
- **Model IV** – 4 categorical parameters + 3 binary parameters + 3 author's parameters = 10 parameters, where the four categorical parameters are the annotation labels (Model I), the three binary parameters represent the message posting options (i.e., first, middle, or last) (Model II), and the three author's parameters are the same as in Model III.

We were interested in the impact of the longer sequences of sentiment transitions on the post's sentiment. To assess this impact, we built the following models:

- **Model V** - 8 categorical parameters. We represented the post by four labels assigned by each annotator to the two previous messages and by four labels assigned by each annotator to the two following messages.
- **Model VI** - 8 categorical parameters + 3 binary parameters = 11 parameters, which combines sentiment labels of Model V and the message posting options within discussion.
- **Model VII** - 8 categorical parameters + 3 author's activity parameters = 11 parameters, which combines sentiment labels of Model V and the authors' activity indicators;
- **Model VIII** - 8 categorical parameters + 3 binary parameters+ 3 author's parameters = 14 parameters. This Model combines all the three characteristics: sentiment labels (Model V), the message posting indicators and the authors' activity indicators.

The following Models use three parameters each. Here, we eschew the sentiment labels. Thus, we remove annotator's bias and only rely on the message posting options and the author's activity indicators
- **Model IX** - 3 binary parameters that represent message's posting options (described in Model II).
- **Model X** - 3 author's parameters, as described in Model III.
- **Model XI** - 3 binary parameters+ 3 author's parameters = 6 parameters.

Next, we assume that the author's activity can affect annotators as well (e.g., they recognize messages of active authors). Thus, the following three Models enhance a possible bias of sentiment labels through author's parameters. We added each of author's parameters to Model I in turn.
- **Model XII** - 4 categorical parameters + 1 binary parameter which shows whether the message author is the first author who started the discussion = 5 parameters.
- **Model XIII** - 4 categorical parameters + 1 parameter which represented the fact that the author just joined the discussion = 5 parameters.
- **Model XIV** - 4 categorical parameters + 1 parameter which represented author's prolificity = 5 parameters.

Table 1 lists the models and the number of parameters each model uses.

**Table 1: ECE models and their parameters.**

| Parameters | | Models | | | | | | | | | | | | | |
|---|---|---|---|---|---|---|---|---|---|---|---|---|---|---|---|
| | | I | II | III | IV | V | VI | VII | VIII | IX | X | XI | XII | XIII | XIV |
| Sentiment labels | categorical | 4 | 4 | 4 | 4 | 8 | 8 | 8 | 8 | | | | 4 | 4 | 4 |
| Message posting options | binary | | 3 | | 3 | | 3 | | 3 | 3 | | 3 | | | |
| The author activity | binary | | | 3 | 3 | | | 3 | 3 | | 3 | 3 | 1 | 1 | 1 |

### 3.3. Model Evaluation

We apply four settings of multi-class classifications of the IVF data set. In those ML settings, the data is classified into six, five, four, and three sentiment categories respectively; Sec. 4.1 provides the details. In every multi-class classification setting, we apply one model in one multi-class sentiment classification. As a result, we have 14x 4 = 52 supervised Machine Learning (ML) tasks to evaluate performance of the ECE models. We apply the model to all messages in the discussion threads. We then separately use Support Vector Machine (SVM) and Conditional Random Fields (CRF), with 10-fold cross validation, to classify sentiments of every message. Thus, we conduct 104 experiments.

*Precision*(macro) serves as the model prowess criterion. It is the average of *Precisions* calculated for each sentiment category, where all the categories are treated equally:

$$Precision = \frac{\sum_{i=1}^{l} \frac{tp_i}{tp_i + fp_i}}{l}$$

where $tp_i$ indicates the number of correctly recognized messages labeled with the sentiment category $i$, $fp_i$ is the number of messages incorrectly assigned to the sentiment category $i$, and $l$ is the number of sentiment categories. It signifies per-class agreement of the data sentiment labels with those obtained by combination of the ECE model and a classifier.

## 4. Empirical Study

4.1 Data set

We work with data obtained from In Vitro Fertilization (IVF) website[2]. The data have been introduced in (Sokolova and Bobicev, 2013) and is available on demand. We analyze discussions from the IVF Ages 35+ sub-forum[3].
- The data set consists of 1321 messages written by 359 female authors and posted on 80 discussions. The average length of the discussion - 16.5 posts (s.t.d. = 9.6); the average number of participants in one discussion comes to 9-10 persons (s.t.d. = 4).The average post had 750 characters and 5-10 sentences.

---
[2] http://www.ivf.ca/
[3] http://ivf.ca/forums/forum/166-ivf-ages-35/

- Discussants form a relatively narrow group: there are 359 female authors;. The forum participants contributed to 80 topics we were working with. 15 authors we denoted as the most prolific; they posted almost 45% of posts. 172 persons posted only one message each.
- There were 73 first authors in the 80 annotated threads, 5 of which started 2 topics and one started 3. The first author who started the thread was usually rather active in the initiated topic and posted in average 4 messages in the started thread.
- In each thread, in average 9.5 posts were written by the participants who joined the discussion for the first time. These messages formed around 64% of topic's posts.

We constructed four multi-class categorizations of the data set:

**6 classes**. The original data with 1321 messages, with the following annotation: 117 posts annotated with 'confusion', 310 posts annotated with 'encouragement', 162 posts annotated with 'endorsement', 124 posts annotated with 'gratitude', 433 posts annotated with 'facts', 175 posts marked as 'ambiguous'; the majority class F-score = 0.162;

**5 classes**. In the next set we removed all posts marked as 'ambiguous' and experimented with 1146 unambiguous messages, including 117 posts annotated with 'confusion', 310 posts annotated with 'encouragement', 162 posts annotated with 'endorsement', 124 posts annotated with 'gratitude', 433 posts annotated with 'facts'; the majority class F-score = 0.207;

**4 classes**. Next, we decided to minimize the number of labels and merged 'encouragement', 'gratitude' and 'endorsement' in one 'positive' class leaving 'confusion' as 'negative', thus obtaining 596 'positive' posts, 117 'negative' posts, 433 'factual' posts, and 175 ambiguous posts; 1321 in total; the majority class F-score = 0.280;

**3 classes**. Next, we removed all posts marked as 'ambiguous' obtaining the set of 1146 messages which included 596 'positive' posts, 117 'negative' posts, and 433 'factual' posts; the majority class F-score = 0.355.

### 4.2. Empirical assessment of the ECE models

We apply Support Vector Machines (SVM, the logistic model, exhaustive search among normalized poly kernels 1 - 5, soft margin 1 – 5, WEKA toolkit) and Conditional Random Fields (CRF, Maximum Likelihood, Mallet toolkit). SVM has shown a reliable performance in sentiment analysis of social networks. At the same time, we expect CRF to benefit from the feature sets that are sequences of possibly dependent random variables. The best classifiers were found through 10-fold cross-validation.

We provide the results obtained on Bag-of-Words (BOW)/unigram model as our benchmark. The total unigram count was 7787 unique words. Words with occurrence 1 and 2 were mostly out-off-vocabulary; we did not use them in the text representation. As the result, we used 3302 unigrams with occurrence > 3 to represent the messages.

We compare the model performance through ranking. For each classification task, we rank the obtained *Precisions* in the descending order. The ECE model with the lowest rank shows the most reliable performance. **Model IX** provided the best ECE assessment when the data sets were classified by SVM. The model's total rank was 8, i.e., rank 1 in 6-class, 3 – in 5-class, and ranks 2 – in 4-class and 3-class classifications. Model IX has 3 binary parameters. **Models I** and **XII** provided a tied-ranked assessment when the data sets were classified by CRF. Their ranks were 7, i.e., for Model I: rank 2 in the 6-class, 3 – in 5-class, and ranks 1 – in 4-class and 3-class classifications, for Model XII: ranks 1 in 6-class and 5-class, 2 – in 4-class, and rank 3 in 3-class classifications.   Model I has 4 categorical parameters, Model XII has four categorical and 1 binary parameters. Below, Table 2 and Fig. 2 report *Precision* provided by the ECE models in SVM classification,   Table 3 and Fig 3 report the ECE model ranking, based on SVM.  Table 4 and Fig 4 report *Precision* provided by the ECE models in CRF classification, Table 5 and Fig 5 report the ECE model ranking, based on CRF.

**Table 2: Precision of the sentiment classification by SVM; for each problem/row, the highest Precision is in bold.**

|         | *BoW*  | I     | II    | III   | IV    | V     | VI    | VII   | VIII  | IX    | X     | XI    | XII   | XIII  | XIV   |
|---------|--------|-------|-------|-------|-------|-------|-------|-------|-------|-------|-------|-------|-------|-------|-------|
| 6-class | *0.415* | 0.375 | 0.380 | 0.380 | 0.373 | 0.373 | 0.396 | 0.398 | 0.412 | **0.425** | 0.182 | 0.254 | 0.393 | 0.348 | 0.362 |
| 5-class | *0.515* | 0.447 | 0.433 | 0.470 | 0.467 | 0.467 | 0.469 | **0.520** | 0.515 | 0.515 | 0.247 | 0.36  | 0.486 | 0.432 | 0.45  |
| 4-class | *0.485* | 0.495 | **0.506** | 0.472 | 0.485 | 0.485 | 0.473 | 0.472 | 0.481 | 0.503 | 0.282 | 0.364 | 0.495 | 0.468 | 0.48  |
| 3-class | *0.635* | 0.604 | 0.621 | 0.617 | 0.608 | 0.608 | 0.640 | 0.640 | **0.649** | 0.648 | 0.358 | 0.582 | 0.626 | 0.629 | 0.62  |

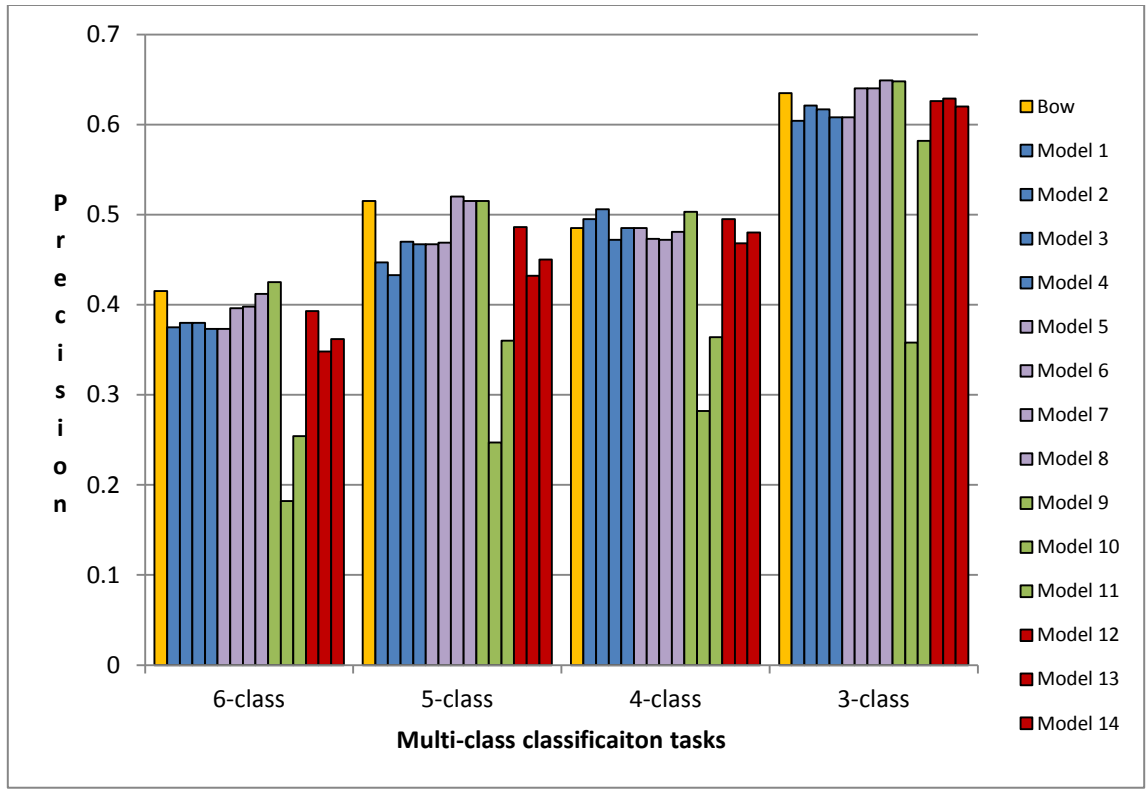

Figure 3: Model performance – Precision, SVM.

**Table 3:   The model rankings for the corresponding tasks – SVM; model IX is the best ranked model.**

|         | *BoW* | I    | II   | III  | IV   | V    | VI   | VII  | VIII | **IX** | X  | XI | XII  | XIII | XIV |
|---------|-------|------|------|------|------|------|------|------|------|--------|----|----|------|------|-----|
| 6-class | *2*   | 9    | 7.5  | 7.5  | 10.5 | 10.5 | 5    | 4    | 3    | **1**  | 15 | 14 | 6    | 13   | 12  |
| 5-class | *3*   | 11   | 12   | 6    | 8.5  | 8.5  | 7    | 1    | 3    | **3**  | 15 | 14 | 5    | 13   | 10  |
| 4-class | *6*   | 3.5  | 1    | 11.5 | 6    | 6    | 10   | 11.5 | 8    | **2**  | 15 | 14 | 3.5  | 13   | 9   |
| 3-class | *5*   | 13   | 8    | 10   | 11.5 | 11.5 | 3.5  | 3.5  | 1    | **2**  | 15 | 14 | 7    | 6    | 9   |

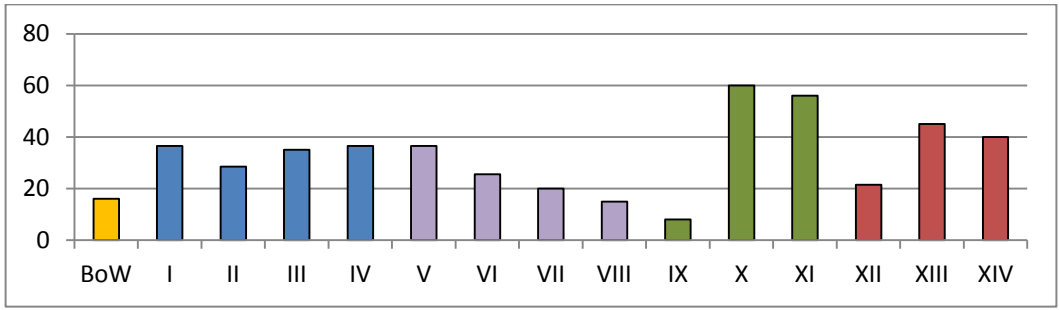

**Figure 4: Total rankings of the models. The lower numbers show a better performance. - SVM.**

**Table 4: Precision of the sentiment classification by CRF; for each problem/row, the highest Precision is in bold.**

|         | BoW   | I     | II    | III   | IV    | V     | VI    | VII   | VIII  | IX    | X     | XI    | XII   | XIII  | XIV   |
|---------|-------|-------|-------|-------|-------|-------|-------|-------|-------|-------|-------|-------|-------|-------|-------|
| 6-class | *0.369* | 0.621 | 0.613 | 0.568 | 0.57  | 0.516 | 0.516 | 0.434 | 0.438 | 0.149 | 0.245 | 0.265 | **0.644** | 0.614 | 0.582 |
| 5-class | *0.441* | 0.619 | 0.616 | 0.601 | 0.582 | 0.580 | 0.566 | 0.507 | 0.507 | 0.230 | 0.334 | 0.321 | **0.672** | 0.643 | 0.593 |
| 4-class | *0.452* | **0.681** | 0.654 | 0.618 | 0.628 | 0.570 | 0.568 | 0.504 | 0.489 | 0.258 | 0.394 | 0.326 | 0.662 | 0.649 | 0.634 |
| 3-class | *0.611* | **0.732** | 0.711 | 0.704 | 0.692 | 0.669 | 0.671 | 0.622 | 0.64  | 0.435 | 0.596 | 0.539 | 0.713 | 0.706 | 0.717 |

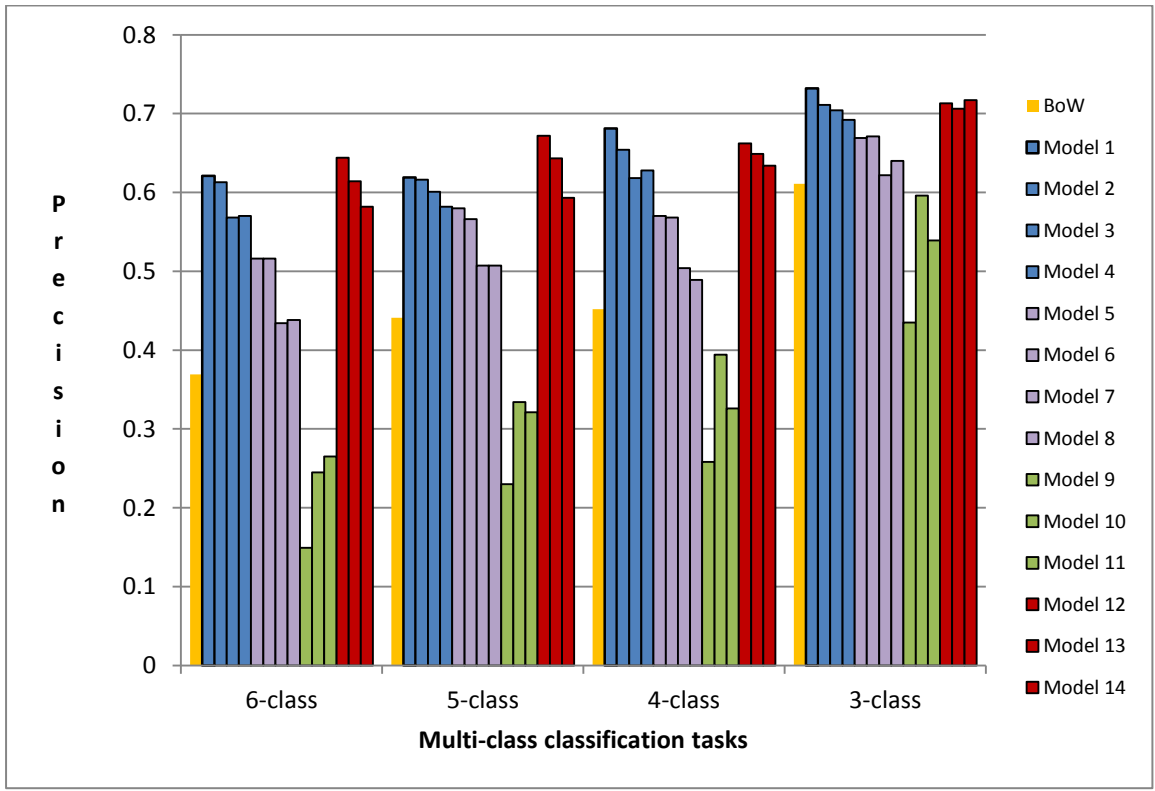

**Figure 5: Model performance – Precision, CRF.**

**Table 5: Rankings of the model performance across the tasks - CRF. models I and XII share the best ranking.**

|         | BoW | I | II | III | IV | V   | VI  | VII  | VIII | IX | X  | XI | XII | XIII | XIV |
|---------|-----|---|----|-----|----|-----|-----|------|------|----|----|----|-----|------|-----|
| 6-class | 12  | 2 | 4  | 7   | 6  | 8.5 | 8.5 | 11   | 10   | 15 | 14 | 13 | 1   | 3    | 5   |
| 5-class | 12  | 3 | 4  | 5   | 7  | 8   | 9   | 10.5 | 10.5 | 15 | 13 | 14 | 1   | 2    | 6   |
| 4-class | 12  | 1 | 3  | 7   | 6  | 8   | 9   | 10   | 11   | 15 | 13 | 14 | 2   | 4    | 5   |
| 3-class | 12  | 1 | 4  | 6   | 7  | 9   | 8   | 11   | 10   | 15 | 13 | 14 | 3   | 5    | 2   |

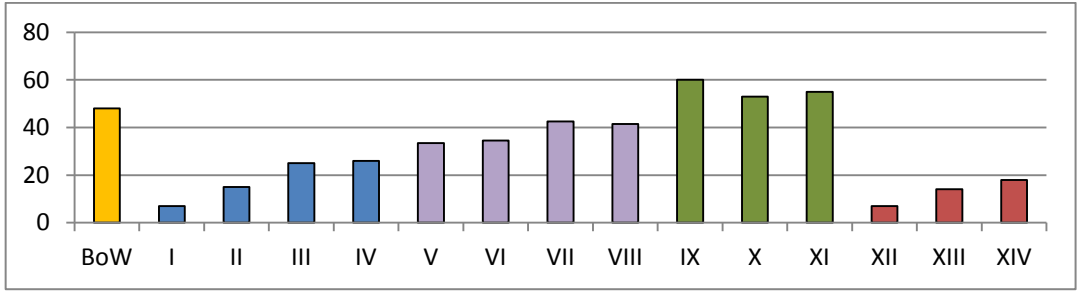

**Figure 6: Total Rankings of the models. The lower numbers show a better performance. – CRF.**

## 5. Discussion

Our ECE analysis is based in three factors: *who* expressed the message sentiments (the authors and their activity), *why* those sentiments appear in the message (sentiments of adjoined messages as perceived by readers), and *where* the sentiment-bearing message appears (the message's posting option in the discussion). Whereas many sentiment analytics studies rely on textual features (Poria et al, 2020), some studies eschew them, e.g., Liu et al (2017). Our models do not use the textual content of the messages. This lexicon-independent approach makes the models suitable for the online environment where lexical aspects significantly depend on the sociodemographic profile of the message author (Hilte et al, 2020).

Selection of performance evaluation metrics is an essential part of Machine Learning applications, where measures can enhance the study impact or diminish it (Flach, 2019). We use *Precision* as the main evaluation measure. Whereas *Recall* captures the ratio of correctly found sentiments for each sentiment category, *Precision* captures the ratio of sentiments that were correctly identified for every sentiment category. A higher model's *Precision* means that the model identifies the target sentiment with fewer errors, making *Precision* more useful than *Recall* when we assess ECE. We provide *Precision*, *Recall*, and *F-score* results in Appendix.

We compare sentiment classification results provided by multiple models (14 models and a benchmark model) on four data sets. Omnibus statistical tests that compare performance of multiple algorithms on multiple data sets, e.g., Friedman test, ANOVA, rely on the number of data sets being larger than the number of the classifiers (García et al, 2010; Stapor, 2017;). Thus, we provide a relative comparison of the ECE models by ranking the classification results obtained with the use of those models. Model IX, with 3 binary parameters of message's posting options, achieved the best ranking in SVM classification. Models I, 4 categorical sentiment parameters, and Model XII, 4 categorical sentiment parameters and 1 binary parameter of the author's activity, share the best ranking in CRF classification.

We have focused on sentiments shared among the discussion participants. At the same time, sharing of sentiments is only a part of Echo-Chamber Effect. Selective usage of information and information omission also contribute to it. Avoidance of opposing information can be achieved through physical distancing, inattention, biased interpretation, forgetting, and self-handicapping (Golman et al, 2017). Investigation of physical avoidance, forgetting, and self-handicapping may be unfeasible: to study those phenomena, e.g., who does not participate in the forum and their reasons for doing so, we have to conduct extensive and rather expensive surveys. At the same time, inattention and biased interpretation can be addressed by extending our analysis to the textual message content and applying advanced semantic and pragmatic tools to it.

## 6. Conclusions and Future Work

We assess Echo-Chamber Effect by applying a supervised multi-class sentiment analysis. We list the novelty factors of our study: i) it extends the ECE analysis to the online forums, thus going beyond the Twitter data; ii) we access the ECE models under four choices of multi-class sentiment classification, thus providing evaluation for every measure under different sentiment settings; iii) our study contributes to expansion of the ECE analysis in the domain of online medical discussions.

We proposed ECE assessment in online forums that relies in readers' perception of the expressed sentiments. Using data sets collected from a medical forum, we compared performance of SVM-based and CRF-based classifiers on 14 ECE models. We empirically evaluated those models in 6-,5-,4-, and 3-class classification tasks. Our empirical results show that ECE can be reliably evaluated without a tedious lexical study of the message content. The best obtained *Precision*, with benchmark *Precision* in brackets: in SVM classification - 0.425 for 6 classes (0.415), 0.520 for 5 classes (0.515), 0.506 for 4 classes (0.485), and 0.649 for 3 classes (0.655); in CRF classification - 0.644 for 6 classes (0.369), 0.672 for 5 classes (0.441), 0.681 for 4 classes (0.452), and 0.732 for 3 classes (0.611).

For future work, we plan to collect data from another medical forum and replicate the study on the new data set. Also, we applied a supervised learning approach which relies on access to the data set being fully manually annotated. A vast and ever growing volume of messages posted on social media makes availability of fully annotated data unrealistic. To reduce dependency on manual annotation, we plan to transition to semi-supervised learning. In the future work we plan to address information avoidance and its tactics such as inattention, biased interpretation, and omission by including the textual content analysis in our research.

## Appendix.   F-score, Precision and Recall of the ECE Assessment

We report macro measures *Precision*, *Recall*, and *F-score* obtained in the ECE empirical assessment. All the obtained F-scores outperformed the majority class baselines; refer to Sec 4.1 for the details.   To put the classification results in perspective, we use Bag-of-Words (BOW), i.e. the unigram representation, as our benchmark.   The total unigram count was 7787 unique words. Words with occurrence 1 and 2 were mostly out-off-vocabulary (OOV) words, e.g., misspelling, typos, non-standard abbreviations. We removed OOV words while building BOW.   As the result, our BOW used 3302 unigrams with occurrence > 3 to represent the messages.  Table A1 reports the classification results. Note that SVM results on BOW significantly outperform CRF (P = 0.0031).

| Sets | # | 6-class | | | 5-class | | | 4-class | | | 3-class | | |
|---|---|---|---|---|---|---|---|---|---|---|---|---|---|
| | | P | R | F | P | R | F | P | R | F | P | R | F |
| SVM | | | | | | | | | | | | | |
| BOW | 3302 | 0.415 | 0.413 | 0.404 | 0.515 | 0.510 | 0.501 | 0.485 | 0.497 | 0.483 | 0.635 | 0.644 | 0.634 |
| CRF | | | | | | | | | | | | | |
| BOW | 3302 | 0.369 | 0.307 | 0.333 | 0.441 | 0.393 | 0.415 | 0.452 | 0.415 | 0.432 | 0.611 | 0.544 | 0.575 |

**Table A1:**   The benchmark results obtained on the unigram representation of the data.

For the SVM-based classification, the best F-scores slightly outperformed BOW's F-score for 6,5,4 classes, and came almost equal for 3 classes (0.633 vs 0.634).  For CRF-based classification, BOW's results were significantly lower than those of SVM; CFR's best results considerably outperformed those obtained on BOW's representation.   Tables A2-A7 report the results obtained on the models introduced in Section 3. Tables A2 - A4 report SVM's performance, and Tables  A5- A7  – on CRF.

For SVM, the best F-score is consistently obtained when the model parameters include sentiment labels, the author activity information and the information about post's position in the discussion (Model VIII). For 6-class, 5-class and 4-class classification, the best F-score were achieved when all the parameters were used, for 3-class classification the best F-score is obtained without post's position parameters (Model VII).   Leaving out sentiment labels considerably decreased classification results (Table A3).  The best F-score = 0.253 was computed on the three author's   activity parameters (Model X). Adding the message position parameters (Model XI) had actually decreased the results – F-score =  0.271.

Enhancing sentiment labels with singular indicators of author's activity (Table A4) has shown inconclusive results if we compare them with the sentiment labels enhanced with the three author's activity parameters (Model III). When sentiment features were augmented with the "first author" indicator (Model XII), F-score improved for all the problems. CRF-based classification obtained higher F-score than SVM on all the four classification tasks. However, CRF results show that its performance is highly volatile and depends on messages being represented through the sentiment labels of immediate preceding messages and following messages.

For 6, 5, and 3 classes, the overall best F-score and the 2$^{nd}$ best F-score are achieved either on sentiment labels augmented with the "first author "indicator (Model XII, Table A7) or on the four sentiment labels along (Model I, Table A5). For 4 classes, the best and second best F-score are achieved on sentiment labels augmented with the "first author" and the "newcomer "indicators respectively (Models XII, XIII, Table A7). Representing messages without sentiment labels decreased CFR ability (Table A6). For 6 classes, the best F-score = 0.249 (Model XI) is lower than any other F-score for the classification problem; the same holds for 5, 4, and 3 classes.

| Models | # | 6-class | | | 5-class | | | 4-class | | | 3-class | | |
|---|---|---|---|---|---|---|---|---|---|---|---|---|---|
| | | P | R | F | P | R | F | P | R | F | P | R | F |
| I | 4 | 0.375 | 0.402 | 0.364 | 0.447 | 0.468 | 0.440 | 0.495 | 0.518 | 0.492 | 0.604 | 0.608 | *0.601* |
| II | 7 | 0.380 | 0.401 | *0.361* | 0.433 | 0.456 | *0.425* | 0.506 | 0.522 | 0.491 | 0.621 | 0.613 | 0.604 |
| III | 7 | 0.380 | 0.413 | 0.387 | 0.470 | 0.496 | 0.468 | 0.472 | 0.500 | 0.478 | 0.617 | 0.616 | 0.613 |
| IV | 10 | 0.373 | 0.409 | 0.382 | 0.467 | 0.496 | 0.467 | 0.485 | 0.505 | *0.486* | 0.608 | 0.609 | 0.604 |
| V | 8 | 0.396 | 0.422 | 0.374 | 0.469 | 0.480 | 0.429 | 0.473 | 0.542 | 0.499 | 0.640 | 0.634 | 0.626 |
| VI | 11 | 0.398 | 0.423 | 0.368 | 0.520 | 0.476 | 0.426 | 0.472 | 0.541 | 0.498 | 0.640 | 0.635 | 0.627 |
| VII | 11 | 0.412 | 0.447 | 0.405 | 0.515 | 0.523 | **0.504**[*] | 0.481 | 0.552 | 0.508 | 0.649 | 0.640 | **0.633**[*] |
| VIII | 14 | 0.425 | 0.448 | **0.411**[*] | 0.515 | 0.521 | **0.504***  | 0.503 | 0.561 | **0.518**[*] | 0.648 | 0.640 | 0.631 |

Table A2: Classification results for SVM on Models I - VIII. For each task, the highest measure is in bold; the lowest – in *italic*.[*] indicates SVM's best F-score for the problem or its tie.

| Models | # | 6-class | | | 5-class | | | 4-class | | | 3-class | | |
|---|---|---|---|---|---|---|---|---|---|---|---|---|---|
| | | P | R | F | P | R | F | P | R | F | P | R | F |
| IX | 3 | 0.182 | 0.344 | *0.211*[#] | 0.247 | 0.392 | *0.263*[#] | 0.282 | 0.476 | *0.334*[#] | 0.358 | 0.546 | *0.410*[#] |
| X | 3 | 0.254 | 0.368 | **0.253** | 0.360 | 0.424 | 0.322 | 0.364 | 0.464 | 0.360 | 0.582 | 0.575 | 0.539 |
| XI | 6 | 0.300 | 0.363 | 0.271 | 0.341 | 0.413 | **0.328** | 0.407 | 0.484 | **0.364** | 0.581 | 0.572 | **0.541** |

Table A3: Classification results for SVM on Models IX- XI. For each task, the highest measure is in bold; the lowest – in *italic*. .[#] indicates SVM's worst F-score for the problem.

| Models | # | 6-class | | | 5-class | | | 4-class | | | 3-class | | |
|---|---|---|---|---|---|---|---|---|---|---|---|---|---|
| | | P | R | F | P | R | F | P | R | F | P | R | F |
| XII | 5 | 0.393 | 0.419 | **0.393** | 0.486 | 0.491 | **0.476** | 0.495 | 0.516 | **0.497** | 0.626 | 0.619 | 0.614 |
| XIII | 5 | 0.348 | 0.376 | *0.350* | 0.432 | 0.431 | *0.422* | 0.468 | 0.499 | *0.476* | 0.629 | 0.626 | **0.622** |
| XIV | 5 | 0.362 | 0.397 | 0.356 | 0.450 | 0.462 | 0.433 | 0.480 | 0.508 | 0.483 | 0.620 | 0.617 | *0.610* |

Table A4: Classification results for SVM on Models XII – XIV. For each task, the highest measure is in bold; the lowest – in *italic*

| Models | # | 6-class | | | 5-class | | | 4-class | | | 3-class | | |
|---|---|---|---|---|---|---|---|---|---|---|---|---|---|
| | | P | R | F | P | R | F | P | R | F | P | R | F |
| I | 4 | 0.621 | 0.605 | **0.613** | 0.619 | 0.614 | **0.616** | 0.681 | 0.652 | **0.665*** | 0.732 | 0.704 | **0.717*** |
| II | 7 | 0.613 | 0.602 | 0.607 | 0.616 | 0.610 | 0.613 | 0.654 | 0.635 | 0.644 | 0.711 | 0.711 | 0.711 |
| III | 7 | 0.568 | 0.570 | 0.569 | 0.601 | 0.610 | 0.605 | 0.618 | 0.601 | 0.609 | 0.704 | 0.693 | 0.698 |
| IV | 10 | 0.570 | 0.566 | 0.568 | 0.582 | 0.587 | 0.584 | 0.628 | 0.612 | 0.620 | 0.692 | 0.694 | 0.692 |
| V | 8 | 0.516 | 0.490 | 0.502 | 0.580 | 0.540 | 0.558 | 0.570 | 0.524 | 0.545 | 0.669 | 0.609 | 0.637 |
| VI | 11 | 0.516 | 0.487 | 0.500 | 0.566 | 0.535 | 0.548 | 0.568 | 0.528 | 0.546 | 0.671 | 0.612 | 0.640 |
| VII | 11 | 0.434 | 0.433 | *0.432* | 0.507 | 0.515 | 0.510 | 0.504 | 0.482 | 0.492 | 0.622 | 0.608 | *0.614* |
| VIII | 14 | 0.438 | 0.436 | 0.436 | 0.507 | 0.500 | *0.503* | 0.489 | 0.480 | *0.483* | 0.640 | 0.625 | 0.631 |

Table A5: Classification results for CRF on Models I - VIII. For each task, the highest measure is in bold. the lowest – in *italic.* * indicates CRF's best F-score for the problem.

| Models | # | 6-class | | | 5-class | | | 4-class | | | 3-class | | |
|---|---|---|---|---|---|---|---|---|---|---|---|---|---|
| | | P | R | F | P | R | F | P | R | F | P | R | F |
| IX | 3 | 0.149 | 0.233 | *0.180[#]* | 0.230 | 0.276 | *0.249[#]* | 0.258 | 0.354 | *0.295[#]* | 0.435 | 0.463 | *0.445[#]* |
| X | 3 | 0.245 | 0.245 | 0.243 | 0.334 | 0.314 | **0.315** | 0.394 | 0.350 | 0.369 | 0.596 | 0.498 | **0.540** |
| XI | 6 | 0.265 | 0.238 | **0.249** | 0.321 | 0.287 | 0.302 | 0.326 | 0.350 | 0.382 | 0.539 | 0.466 | 0.498 |

Table A6: Classification results for CRF on Models IX - XI. For each task, the highest measure is in bold. the lowest – in *italic.* .[#] indicates SVM's worst F-score for the problem.

| Models | # | 6-class | | | 5-class | | | 4-class | | | 3-class | | |
|---|---|---|---|---|---|---|---|---|---|---|---|---|---|
| | | P | R | F | P | R | F | P | R | F | P | R | F |
| XII | 5 | 0.644 | 0.641 | **0.642*** | 0.672 | 0.674 | **0.673*** | 0.662 | 0.648 | **0.655** | 0.713 | 0.709 | **0.711** |
| XIII | 5 | 0.614 | 0.607 | 0.610 | 0.643 | 0.643 | 0.642 | 0.649 | 0.630 | 0.639 | 0.706 | 0.699 | *0.702* |
| XIV | 5 | 0.582 | 0.585 | *0.583* | 0.593 | 0.589 | *0.591* | 0.634 | 0.614 | *0.624* | 0.717 | 0.697 | 0.705 |

**Table A7: Classification results for CRF on Models XII - XIV. For each task, the highest measure is in bold; the lowest – in *italic*.*  indicates CRF's best F-score for the problem.**